\pdfoutput=1
\documentclass[11pt]{article}

\usepackage[preprint]{acl}

\usepackage{times}
\usepackage{latexsym}
\usepackage{multirow} 
\usepackage{longtable}

\usepackage[T1]{fontenc}
\usepackage[utf8]{inputenc}

\usepackage{microtype}

\usepackage{inconsolata}

\usepackage{graphicx}
\usepackage{algorithm}
\usepackage{algpseudocode}
\usepackage{booktabs}
\usepackage{url}
\usepackage{natbib}
\title{PromotionGo at SemEval-2025 Task 11: A Feature-Centric Framework for Cross-Lingual Multi-Emotion Detection in Short Texts}

\author{Ziyi Huang \\
  Hubei University \\
  \texttt{ziyihuang@hubu.edu.cn} \\\And
  Xia Cui \\
  Manchester Metropolitan University \\
  \texttt{x.cui@mmu.ac.uk} \\}

\begin{document}
\maketitle
% papre writing instruction : https://github.com/nedjmaou/Writing_a_task_description_paper?tab=readme-ov-file

\begin{abstract}
This paper presents our system for SemEval 2025 Task 11: Bridging the Gap in Text-Based Emotion Detection (Track A)~\citep{muhammad-etal-2025-semeval}, which focuses on multi-label emotion detection in short texts. We propose a feature-centric framework that dynamically adapts document representations and learning algorithms to optimize language-specific performance. Our study evaluates three key components: document representation, dimensionality reduction, and model training in 28 languages, highlighting five for detailed analysis. The results show that TF-IDF remains highly effective for low-resource languages, while contextual embeddings like FastText and transformer-based document representations, such as those produced by Sentence-BERT, exhibit language-specific strengths. Principal Component Analysis (PCA) reduces training time without compromising performance, particularly benefiting FastText and neural models such as Multi-Layer Perceptrons (MLP). Computational efficiency analysis underscores the trade-off between model complexity and processing cost. Our framework provides a scalable solution for multilingual emotion detection, addressing the challenges of linguistic diversity and resource constraints.

% This paper presents our system developed for the SemEval 2025 Task 11: Bridging the Gap in Text-Based Emotion Detection Track A: Multi-label Emotion Detection. This task aims at predicting the perceived emotion(s) of a speaker according to a given target text snippet, more specifically to predict multiple emotions. Our system addresses this challenge by employing a multilabel classification approach that enables the system to predict multiple emotions simultaneously. Moreover, we

\end{abstract}

\section{Introduction}
% (Why need to label, what is multi-label, the challenge of multi-label comparing to single)

Emotion labeling in Natural Language Processing (NLP) is critical for enabling machines to better interpret human emotional expressions, fostering empathetic and context-aware AI systems. Traditional single-label emotion detection oversimplifies human affect by assigning a single dominant emotion to text, ignoring the complex spectrum of overlapping emotions often present in real-world scenarios~\citep{plutchik2001nature}. In contrast, multi-label emotion detection aligns more closely with authentic human experiences, where texts may simultaneously express multiple emotions (e.g., joy and surprise, or sadness and anger). It provides ecologically valid representations of emotional complexity, better reflecting nuanced psychological states. However, multi-label emotion detection can be challenging, such as models must account for emotion co-occurrence, resolve subtle semantic ambiguities, and avoid overfitting to sparse or imbalanced label distributions~\citep{ekman1992argument,deng2020multi}.

% The SemEval 2025 Task 11 (cite) is a shared task designed to advance the field of computational emotion analysis by addressing the multifaceted nature of emotion expression in text. The task encompasses 28 languages, including English, German, Russian, and Ukrainian. Track A, which focuses on Multi-label Emotion Detection, recognizes the phenomenon that text in any language may concurrently convey multiple emotions.

Traditional approaches to multi-label emotion detection have predominantly relied on \textit{feature-centric frameworks} that leverage handcrafted linguistic and statistical features.
While effective in monolingual settings, such frameworks often required language-specific resources (e.g., lexicons for each target language~\citep{baccianella2010sentiwordnet}), limiting cross-lingual scalability.
These features (e.g., lexicons, syntactic patterns) often fail to model the complex interdependencies and contextual nuances required for multi-label emotion detection, as they struggle to capture dynamic label correlations and contextualized affective semantics~\citep{baccianella2010sentiwordnet,mohammad2018semeval,bostan2018good}. 

In this paper, we conduct a comprehensive study on the development of a feature-centric framework to address the challenges of cross-lingual adaptability and multi-label emotion detection through a three-stage methodological pipeline: (1) feature extraction/document representation, (2) dimensionality reduction and (3) model training.
For (1), we unify diverse feature representations ranging from interpretable shallow features (e.g., TF-IDF, Bag-of-Words) to contextually rich embeddings (e.g., FastText, BPE) and Transformer-based semantic encodings (e.g., Sentence-BERT). 
For (2), we reduce the dimensionality of document representations to prevent model overfitting and accelerate the subsequent step.
For (3), we systematically evaluate traditional machine learning classifiers (e.g., SVM, RF) and deep learning architectures (e.g., MLP) to optimize label dependency modeling. The modular design allows interchangeable classifier integration, balancing interpretability (via traditional models) and performance (via neural approaches) for diverse multilingual use cases.

The source code for this paper is publicly available on GitHub\footnote{\url{https://github.com/YhzyY/SemEval2025-Task11}}.

% \section{Background}

% \emph{In your own words, summarize important details about the task setup: kind of input and output (give an example if possible); what datasets were used, including language, genre, and size. If there were multiple tracks, say which you participated in. \\ 
% Here or in other sections, cite related work that will help the reader to understand your contribution and what aspects of it are novel.\\}

% related work (todo):
% Sechidis K., Tsoumakas G., Vlahavas I. (2011) On the Stratification of Multi-Label Data. In: Gunopulos D., Hofmann T., Malerba D., Vazirgiannis M. (eds) Machine Learning and Knowledge Discovery in Databases. ECML PKDD 2011. Lecture Notes in Computer Science, vol 6913. Springer, Berlin, Heidelberg.

\section{Background and System Overview}

% \emph{Key algorithms and modeling decisions in your system; resources used beyond the provided training data; challenging aspects of the task and how your system addresses them. This may require multiple pages and several subsections, and should allow the reader to mostly reimplement your system’s algorithms.\\
% Use equations and pseudocode if they help convey your original design decisions, as well as explaining them in English. If you are using a widely popular model/algorithm like logistic regression, an LSTM, or stochastic gradient descent, a citation will suffice—you do not need to spell out all the mathematical details.\\ 
% Give an example if possible to describe concretely the stages of your algorithm. \\
% If you have multiple systems/configurations, delineate them clearly.\\
% This is likely to be the longest section of your paper.}

% This system is developed on the training set and subsequently validated using the test set. No external datasets were used to boost performance. 
The development pipeline within the system comprised three distinct stages: (1) feature representation utilizing a diverse set of techniques such as TF-IDF, FastText and Sentence-Transformers, (2) dimensionality reduction of feature vectors via Principal Component Analysis (PCA), and (3) model training and prediction employing a suite of algorithms such as Decision Trees (DT) and Multi-Layer Perceptrons (MLP).

\subsection{Document Representation}
Raw text data in human languages are sequences of variable-length symbolic representations, which challenge machine learning algorithms that require fixed-size numeric vectors~\citep{mikolov2013distributed}. An effective document representation is essential for optimal NLP performance. To address the complexities of multi-language data, various feature representation techniques have been explored.

\subsubsection{Traditional Features}
The traditional approaches represent a document $d$ by creating a list of unique words and assigning each word $w$ a numeric value, such as Bag of words (BoW) and Term Frequency-Inverse Document Frequency (TF-IDF).

We use scikit-learn's \texttt{CountVectorizer} to extract BoW features, which represent word occurrence in a document without considering word frequency or significance. This approach considers all words equally, including common terms like “\textit{the}” or “\textit{a}”, which carry minimal meaningful information. In contrast, TF-IDF adjusts for the importance of words by considering both their frequency in a document (TF) and their rarity across the corpus (IDF), reducing the impact of common words. We use scikit-learn’s \texttt{TfidfVectorizer}, where IDF is computed as:
\begin{equation}
\mathrm{IDF}(w, d) = \log\frac{1 + N}{1 + \mathrm{DF}(w)} + 1
\end{equation}
where $N$ is the total number of documents and DF($w$) is the document frequency of $w$.
%\footnote{\url{https://scikit-learn.org/stable/modules/generated/sklearn.feature_extraction.text.CountVectorizer.html}}
%\footnote{\url{https://scikit-learn.org/stable/modules/generated/sklearn.feature_extraction.text.TfidfVectorizer}}

Preprocessing is critical in traditional feature representations, since it directly influences feature quality and model training. To adapt our model to multilingual scenarios and capture nuanced emotional expressions, we employ GemmaTokenizer~\citep{gemmamultilingualwebsite} in the tokenization step, instantiated with the preset "\textit{gemma\_2b\_en}". Trained on a diverse multilingual corpus, GemmaTokenizer excels in handling linguistic diversity and demonstrates robust performance in multilingual tokenization tasks.

\subsubsection{Pretrained Word Embeddings}
Traditional methods like TF-IDF and BoW treat words as discrete, independent units, thereby completely discarding word order and local contextual information during the encoding process, leading to high-dimensional, sparse representations that limit computational efficiency and NLP task performance~\citep{mikolov2013distributed, pennington2014glove}. In addition, their heavy reliance on training corpus results in their inability to handle out-of-vocabulary (OOV) words, as they lack a mechanism to infer the representation or meaning of terms that are not present in the training corpus.

Pre-trained word embeddings offer substantial advantages in capturing semantic meaning, contextual dependencies, and generalization capabilities. Unlike traditional sparse representations, these dense vector embeddings encode rich linguistic features by leveraging large-scale corpora during pretraining, enabling them to model nuanced semantic relationships and syntactic patterns. FastText~\citep{bojanowski2017enrichingwordvectorssubword} can capture both syntactic and semantic relationships by effectively modeling morphological structures. Byte Pair Embeddings (BPEs)\citep{heinzerling-strube-2018-bpemb} decompose words into subwords, while Contextual String Embeddings (CSEs)\citep{akbik2018contextual} provide context-sensitive representations, dynamically adapting to word meanings. We use the Flair NLP Toolkit\footnote{\url{https://github.com/flairNLP/flair}} to extract these embeddings and \texttt{DocumentPoolEmbeddings} for aggregating word-level embeddings into document-level representations via mean pooling.

Given that the cross-lingual representation ability of most mainstream pre-trained language models remains constrained by the limited coverage of training corpora, these models often manifest systematic representational failures when processing low-resource languages excluded from training data. This closed-corpus modeling paradigm inherently imposes significant capabilities limitations in multilingual scenarios. To address this, we leverage Large Language Models (LLMs) to assist with unseen languages by leveraging language family classification. Considering a low-resource language such as the Oromo language, it is not included in the pre-trained FastText embeddings, as outlined in the FastText documentation~\citep{bojanowski2017enrichingwordvectorssubword}. we use Baidu \texttt{Qianfan}\footnote{\url{https://qianfan.readthedocs.io/en/stable/qianfan.html}} model to identify the most linguistically similar supported language in FastText. The text in unseen languages, such as Oromo, is then represented using the embeddings of the identified language.
The full query is presented in Appendix~\ref{sec:appendix-land-class}.
This end-to-end self-adaptive multilingual emotion detection framework significantly enhances the system's ability to process unseen languages while fundamentally eliminating the dependencies on manually annotated language family labels and expert-curated linguistic representation rules, thus circumventing the prohibitive costs of human annotation and resolving the acute scarcity of training data and domain experts in endangered languages.

% Given the fact that many pre-trained word embeddings lack comprehensive coverage across all languages, potentially limiting their efficacy in multilingual tasks, we use Large Language Models (LLM) to assist the unseen languages based on language family classification. For instance, Oromo language was not included in pre-trained FastText embeddings\footnote{\url{https://fasttext.cc/docs/en/crawl-vectors.html}}, we use LLM to identify any FastText's supported languages that exhibit the highest linguistic similarity to Oromo. Then, the text with unseen languages such as Oromo is represented by the embeddings of the identified language. 

\subsubsection{Transformers}
Transformer-based document representations, such as those produced by Sentence-BERT~\citep{reimers-2019-sentence-bert, reimers-2020-multilingual-sentence-bert}, leverage the Transformer architecture to selectively focus on semantically relevant segments of text, thereby enhancing feature extraction and improving representational accuracy. Our system embedded the SentenceTransformers \footnote{\url{https://sbert.net/}} with the pretrained "\textit{paraphrase-multilingual-mpnet-base-v2}" model, which supports over 50 languages. Document embeddings are generated using the library's encode function. 

% Additionally, we utilize BERT-base and RoBERTa-base models via the Flair NLP Toolkit's \texttt{TransformerDocumentEmbeddings} class to produce unified sentence embeddings, with configurable parameters for flexible model comparison.

% Transformer-based document representation like Sentence BERT~\citep{reimers-2019-sentence-bert, reimers-2020-multilingual-sentence-bert} leverages Transformer architecture to focus on the most relevant parts of text, enhancing feature extraction and improving accuracy. We use the implementation from SentenceTransformers library\footnote{\url{https://sbert.net/}}. We first load the pretrained Sentence Transformer model "paraphrase-multilingual-mpnet-base-v2", which is a multilingual model trained on parallel data for 50+ languages, and then the document embeddings can be calculated by calling the encode function provided by \texttt{SentenceTransformers}. 
% Our system leverages BERT-base and RoBERTa-base models through the Flair NLP Toolkit’s \texttt{TransformerDocumentEmbeddings} class to generate unified sentence-level embeddings. By enabling configurable parameter adjustments, this framework supports flexible initialization of distinct embeddings, facilitating streamlined comparative analysis across model configurations.
%\footnote{\url{https://flairnlp.github.io/docs/tutorial-embeddings/transformer-embeddings}}

\subsection{Dimensionality Reduction}
The inherent high cardinality of lexical features within textual data frequently results in high-dimensional embedding, which can lead to computational challenges and model overfitting.
To reduce the dimensionality of document representations, we first normalize the text to unit norm using scikit-learn's \texttt{Normalizer}. Subsequently, Principal component analysis (PCA)~\citep{pearson1901liii} is applied to project the features into a lower-dimensional space. Both the \texttt{Normalizer} and PCA utilize default parameter settings.

%\footnote{\url{https://scikit-learn.org/stable/modules/generated/sklearn.preprocessing.Normalizer.html}}
%\footnote{\url{https://scikit-learn.org/stable/modules/generated/sklearn.decomposition.PCA.html}}

% Principal component analysis (PCA) is a linear dimensionality reduction technique widely used in data analysis, noise filtering, and feature extraction. To reduce the dimension of document embeddings, scikit-learn's \texttt{Normalizer}\footnote{\url{https://scikit-learn.org/stable/modules/generated/sklearn.preprocessing.Normalizer.html}} is used to normalize text to unit norm, and then \texttt{PCA}\footnote{\url{https://scikit-learn.org/stable/modules/generated/sklearn.decomposition.PCA.html}} projects feature to a lower dimensional space. Both the \texttt{Normalizer} and \texttt{PCA} use default parameter settings, which means that the number of components retained after \texttt{PCA} is:

% \begin{equation}
% \mathrm{n\_components} = \mathrm{min}(\mathrm{n\_samples}, \mathrm{n\_features})
% \end{equation}

\subsection{Model Training}
In this section, we present the methods employed for model training, encompassing traditional machine learning approaches as well as simple deep learning architectures such as MLP.
% By feeding extracted document embeddings to learning algorithms, the algorithms can learn patterns and relationships within those language embeddings, therefore recognize and classify emotional expressions.

\subsubsection{Traditional Machine Learning}
As candidate models for training, we employ a variety of traditional machine learning algorithms, including Decision Trees (DT), k-Nearest Neighbors (KNN), Random Forest (RF) and Support Vector Machines (SVM). However, relying solely on these traditional methods often results in suboptimal accuracy~\citep{le2014distributed}. To overcome this limitation, we adopt ensemble learning techniques, specifically constructing a majority voting classifier that aggregates predictions from a collection of base classifiers to improve overall performance.

% Traditional machine learning algorithms are fundamentally statistical or mathematical models that learn patterns from data. 
% Multiple traditional machine learning algorithms are implemented in the system to analyse their performance on multilingual emotion analysis, such as Decision Tree, KNN and SVM.
%\footnote{\url{https://scikit-learn.org/stable/modules/generated/sklearn.tree.DecisionTreeClassifier.html}}
%\footnote{\url{https://scikit-learn.org/stable/modules/generated/sklearn.neighbors.KNeighborsClassifier.html}}
%\footnote{\url{https://scikit-learn.org/stable/modules/generated/sklearn.svm.SVC.html\#sklearn.svm.SVC}}

% the predictions of multiple models and leverages the collective knowledge of them, thus achieve better results than any single model could on its own.
%\texttt{Voting}\footnote{\url{https://scikit-learn.org/stable/modules/generated/sklearn.ensemble.VotingClassifier.html}}

\subsubsection{Deep Learning}
% Employing scikit-learn's \texttt{MLPClassifier}\footnote{\url{https://scikit-learn.org/stable/modules/generated/sklearn.neural_network.MLPClassifier.html}}, 
Multi-Layer Perceptrons (MLPs), with their multi-layered neuron architecture, can learn complex patterns in data, making them well-suited for emotion detection tasks where sentiment often depends on intricate word combinations~\citep{goodfellow2016deep}. 
% Additionally, the flexibility of MLPs, allowing adjustments to hidden layers and neuron counts, enables fine-tuning for specific sentiment analysis tasks and datasets.
% Multi-Layer Perceptron (MLP)'s multi-layered neurons architecture, possesses the capacity to learn complex patterns in data, thus suitable for emotion detection task as sentiment often exhibits complex dependencies on word combination. Furthermore, the inherent flexibility of MLP architectures, allowing for the adjustment of hidden layer configurations and neuron counts, facilitates it fine-tuning for specific sentiment analysis tasks and datasets. 
To account for linguistic variations across languages, we use Grid Search to evaluate multiple parameter combinations and select the one that maximizes the \textit{F1-macro} score, ensuring robust and accurate emotion predictions across diverse linguistic contexts.

% Considering that different languages may have different linguistic characteristics, we use \texttt{GridSearchCV}\footnote{\url{https://scikit-learn.org/stable/modules/generated/sklearn.model_selection.GridSearchCV.html}} to evaluate multiple parameter combinations and select the one that maximize the set evaluation score (which is set to "f1\_macro" in our system according to the task evaluation criterion), leading to more robust and accurate emotion predictions across diverse linguistic contexts.
\section{Experimental Data}

% \emph{How data splits (train/dev/test) are used.\\
% Key details about preprocessing, hyperparameter tuning, etc. that a reader would need to know to replicate your experiments. If space is limited, some of the details can go in an Appendix.\\
% External tools/libraries used, preferably with version number and URL in a footnote.\\
% Summarize the evaluation measures used in the task.\\
% You do not need to devote much—if any—space to discussing the organization of your code or file formats.}

% \subsection{Dataset}

We use the BRIGHTER dataset~\citep{muhammad2025brighterbridginggaphumanannotated,belay-etal-2025-evaluating} provided by SemEval 2025 Task 11 Track A to conduct our experiments., It consists of human-annotated short texts in 28 languages, such as English, German and Russian. The training dataset comprises 65,098 multi-label samples, each annotated with emotion labels — anger, fear, joy, sadness, surprise, and disgust — representing the emotions most likely experienced by the speaker, as inferred from the text.
We used only the provided datasets during the development and evaluation phases, no additional training data was introduced to boost the performance. Table~\ref{training-data statistics} shows the statistics in selected languages, with a full list available in the Appendix~\ref{sec:appendix-data-statistics}.

\begin{table}
\caption{Data Splits: Number of Train (\#train), Development (\#dev) and Test (\#test) samples in our experiments.}
\label{training-data statistics}
\centering
\resizebox{0.7\columnwidth}{!}{%
\begin{tabular}{cccccc}
\toprule
 Language & \#train  & \#dev & \#test \\
\midrule
Marathi & 2415 & 100 & 1000\\
Spanish & 1996 & 184 & 1695\\
Hindi & 2556 & 100 & 1010\\
Romanian & 1241 & 123 & 1119\\
Russian & 2679 & 199 & 1000 \\
\bottomrule
\end{tabular}%
}
\end{table}

% \subsection{Hyperparameter}
% MLPClassifier
\section{Results}

% \emph{Main quantitative findings: How well did your system perform at the task according to official metrics? How does it rank in the competition?\\
% Quantitative analysis: Ablations or other comparisons of different design decisions to better understand what works best. Indicate which data split is used for the analyses (e.g. in table captions). If you modify your system subsequent to the official submission, clearly indicate which results are from the modified system.\\
% Error analysis: Look at some of your system predictions to get a feel for the kinds of mistakes it makes. If appropriate to the task, consider including a confusion matrix or other analysis of error subtypes—you may need to manually tag a small sample for this.}
% \emph{A few summary sentences about your system, results, and ideas for future work.}

We perform experiments in 28 languages and evaluate model performance using the \textit{F1-macro} score. Additionally, we record time consumption and generate confusion matrices to further analyze the models' performance.
%The prediction results for all languages, evaluated under various representation methods and classifiers, are provided in Appendix~\ref{sec:appendix-prediction-accuracy}.

\subsection{Representation Selection}
To investigate the impact of document representation methods on prediction outcomes, we conducted a controlled experiment with varying representations and a fixed learning algorithm.

We experimented with various representation methods, and Table \ref{tab:result-representation} presents the \textit{F1-macro} scores for the best-performing candidates from three representation approaches (i.e. traditional features, pre-trained word embeddings and transformers) across five languages\footnote{Due to the page limit, we randomly selected five languages as examples.}. To reduce the variance introduced by individual classifiers, we employ a majority voting classifier, thereby providing a more stable basis for evaluating the performance differences across various representation methods. The results of all 28 languages can be found in Appendix Section~\ref{sec:appendix-prediction-accuracy}.
% across several languages using different document representations, while the Decision Tree algorithm remains consistent. 
TF-IDF consistently outperforms other document representations across 3 out of 5 selected languages, achieving the highest F1-macro score, particularly in Marathi (0.7438). This highlights TF-IDF’s effectiveness, especially in low-resource languages, as it relies on word frequency rather than pre-trained embeddings. Sentence-BERT (SBERT) shows mixed performance across languages. While it achieves the best result for Romanian (0.5630) and Hindi (0.5682), it performs worse than TF-IDF in the majority of other languages. This suggests that semantically rich contextual embeddings, such as those produced by SBERT, can offer advantages in certain linguistic contexts, but may not consistently outperform simpler lexical representations across all languages. FastText performs moderately well in some cases, such as Romanian (0.4486), but struggles in others, indicating its sensitivity to language-specific characteristics. These results emphasize that while pre-trained embeddings offer advantages in certain contexts, traditional frequency-based representations like TF-IDF remain highly competitive for emotion detection in multilingual settings.

% TF-IDF shows superior overall performance in these languages, 
% achieving the highest scores in every linguistic scenario, 
% especially Marathi's 0.6792 peak score. This suggests that TF-IDF remains highly competitive despite the advancements in embedding techniques, and is effective in low-resource languages since it does not rely on pre-trained word embeddings but is directly based on word frequency. In contrast, Byte Pair Encoding (BPE) exhibits generally poor performance, which could be ascribed to its concentration on subword units, thereby leading to an inadequate representation of semantic or emotional content in short texts, resulting its restricted practicality in short text scenarios. Both FastText and Contextual String Embeddings (CSE) demonstrate language-specific effectiveness, which highlight their sensitivity to linguistic characteristics and insufficient generalization ability, therefore need to be optimized for specific languages when used, and cannot achieve good prediction results for lower-resource language.

% To investigate the impact of document representation methodologies on prediction outcomes, we conducted a controlled experiment where various representation techniques are applied while the learning algorithm is consistently fixed.

% Table \ref{tab:result-representation} shows the system accuracy on several languages when employing diverse document representation methodologies while maintaining a consistent Decision Tree learning algorithm.

\begin{table}
\caption{F1-macro scores for document representations across selected languages using voting classifier.}
\label{tab:result-representation}
\centering
\resizebox{0.85\columnwidth}{!}{%
\begin{tabular}{cccc}
\toprule
Language & TF-IDF & FastText & SBERT \\
\midrule
Marathi & \textbf{0.7438} & 0.4277 & 0.6654 \\
Spanish &\textbf{ 0.6561} & 0.4046 &  0.5867\\
Hindi & 0.4927  & 0.3067 & \textbf{0.5682}\\
Romanian & 0.4358 & 0.4486 & \textbf{0.5630} \\
Russian & \textbf{0.7107} & 0.3472 &  0.5767\\
\bottomrule
\end{tabular}%
}
\end{table}

\subsection{Learning Algorithms}
% By maintaining a consistent document representation methodology, we can analyze the influence of varying learning algorithms on prediction outcomes. 
Using a consistent document representation, we evaluate the impact of different learning algorithms on prediction outcomes. Table \ref{tab:result-learningalgorithms} shows the performance of various algorithms with SBERT embeddings. 
% DT outperforms any other traditional machine learning algorithms.
The performance hierarchy is consistent across languages: MLP > Voting > DT, with MLP demonstrating the best ability to capture complex emotional patterns. The Voting classifier, combining KNN, DT and RF performs moderately, outperforming DT but lagging behind MLP. DT show weaker performance, indicating their limited capacity to model emotional nuances. These results highlight the importance of algorithm choice, with MLP being particularly effective for multilingual emotion detection.
% By using a consistent document representation, we evaluate the impact of different learning algorithms on prediction outcomes. Table \ref{tab:result-learningalgorithms} presents the performance of various learning algorithms using FastText as document representation. The performance hierarchy remains consistent across languages, following MLP > Voting > Decision Tree (DT) > Random Forest (RF), demonstrating MLP's superior capacity to model complex linguistic patterns inherent in emotional analysis. The Voting, which combined KNeighbors, DT, and RF through soft voting and designed to leverage the strengths of these classifiers, exhibits moderate performance, generally surpassing DT and RF but falling short of MLP's efficacy. DT and RF show relatively poor performance, suggesting their insufficient model complexity for capturing nuanced emotional patterns across languages. These findings underscore the importance of algorithm selection in multilingual emotion detection, highlighting MLP as a particularly robust choice when employing those document embeddings.

\begin{table}
\caption{F1-macro scores using SBERT embeddings.}
\label{tab:result-learningalgorithms}
\centering
\resizebox{0.76\columnwidth}{!}{%
\begin{tabular}{cccc}
\toprule
Language & DT & Voting & MLP \\
\midrule
Marathi & 0.4275 & 0.6654 & \textbf{0.8389} \\
Spanish &  0.5022 & 0.5867 & \textbf{0.7076}\\
Hindi & 0.4490  & 0.5682 & \textbf{0.7374}\\
Romanian & 0.5167  & 0.5630 & \textbf{0.6375}\\
Russian & 0.4661 & 0.5767 & \textbf{0.7188}\\
\bottomrule
\end{tabular}%
}
\end{table}

\subsection{Ablation Study}
To access the contribution of individual components, we conducted an ablation study by removing specific modules and analyzing their impact on performance. The only removable component in our system is the dimensionality reduction step. Using the Spanish language dataset as an example, we remove PCA to evaluate its impact on predictive performance.
PCA impacts multilingual emotion detection frameworks differently based on representation-classifier pairings: for TF-IDF, it reduces MLP training time by lowering dimensionality but increases overhead for DT and the Voting classifier without accuracy gains. FastText benefits from PCA-driven noise reduction (i.e. improving accuracy) but incurs higher computational costs to retain variance, whereas SBERT’s performance slightly declines as PCA strips contextual nuances critical for emotion differentiation, despite longer training times. Tree-based models such as DT remain unaffected, prioritizing raw feature hierarchies over reduced embeddings. These results emphasize that PCA’s value depends on representation type (i.e. contextual vs. static) and classifier architecture, advocating for selective use to optimize multilingual systems.
% It streamlining non-contextual features (e.g., TF-IDF, FastText) while preserving Transformer semantics
% The training time without PCA implemented is shown in Table \ref{tab:result-ablation-time-without-pca}, and the training time with PCA implemented is shown in Table \ref{tab:result-ablation-time-with-pca}.

\begin{table}[t]
\caption{Training time in seconds with (w/) and without (w/o) PCA. The best results are bolded.}
\label{tab:result-ablation-time-pca}
\centering
\resizebox{0.99\columnwidth}{!}{
\begin{tabular}{c|cccc}
\toprule
& & DT & Voting & MLP \\
\midrule
~ & TF-IDF & \textbf{1.4894} &  \textbf{59.7822} &  200.7861\\
w/o PCA &FastText & \textbf{0.8623} & \textbf{17.2231} & \textbf{34.3866}\\
~ & SBERT &\textbf{2.6267 } &\textbf{36.9725 } & \textbf{41.6211}\\
\midrule
~ & TF-IDF & 10.7575 & 201.4630 & \textbf{114.9862}\\
w/ PCA & FastText & 0.9941 & 23.4849 & 41.7187\\
~ & SBERT & 2.7614 & 49.2546 & 47.1017\\
\bottomrule
\end{tabular}%
}
\end{table}

\begin{table}[t!]
\caption{F1-macro scores w/ and w/o PCA.}
\label{tab:result-ablation-score-pca}
\centering
\resizebox{0.86\columnwidth}{!}{
\begin{tabular}{c|cccc}
\toprule
& & DT & Voting & MLP \\
\midrule
~ & TF-IDF & \textbf{0.5516} & \textbf{0.6561} & \textbf{0.6284} \\
w/o PCA & FastText & \textbf{0.3825}  & 0.4046 & 0.6479\\
~ & SBERT &  \textbf{0.5022} & \textbf{0.5867} &  0.7076\\
\midrule
~ & TF-IDF & 0.3710  & 0.3931 & 0.6100\\
w/ PCA & FastText & 0.3805 & \textbf{0.4407} & \textbf{0.6558}\\
~ & SBERT & 0.4237 & 0.5038 & \textbf{0.7093} \\
\bottomrule
\end{tabular}%
}
\end{table}

\subsection{Data Imbalance} 
Data imbalance in the training dataset can significantly impact model performance, causing bias towards the majority class and resulting in poor predictions for the minority class. 

% Figure \ref{fig:result-imbalance-hin} and Table \ref{tab:result-imbalance-hin} demonstrate the substantial impact.
For the Hindi dataset trained with FastText and MLP, over 78\% of 2,556 samples are labeled as not "\textit{anger}" leading to a high specificity score (0.9405) but a low recall score (0.5625) for the "\textit{anger}" label, as shown in Table \ref{tab:result-imbalance-hin}. A similar pattern is observed for other emotion labels, highlighting that imbalance between positive and negative samples can undermine prediction accuracy. These results emphasize the significant impact of label distribution on system performance.
% For the Hindi dataset trained on FastText and MLP, out of 2,556 samples, more than 2,000 of them are labeled as "anger negative", leading to a high specificity score 0.9405 but a notably low recall score 0.5625 for the "anger" label, as reflected in Table \ref{tab:result-imbalance-hin}. A similar trend is also observed for other emotion labels, which indicates the imbalance between positive and negative samples can result in relatively low prediction accuracy. These statistics highlight the significant influence of label distribution imbalance on system performance. 

% \begin{figure}
%     \centering
%     \includegraphics[width=1\linewidth]{figure/hin-emotion-distribution.png}
%     \caption{Label Quantity Distribution of Hindi.}
%     \label{fig:result-imbalance-hin}
% \end{figure}

\begin{table}
\caption{Confusion matrix on Hindi language subset.}
\resizebox{\columnwidth}{!}{
\label{tab:result-imbalance-hin}
\centering
\begin{tabular}{ccccccc}
\toprule
 & anger & disgust & fear & joy & sadness & surprise \\
\midrule
TP& 0.5625 & 0.5000 & 0.6429 & 0.6364 & 0.2941 & 0.6667\\
TN& 0.9405 & 1.0000 & 0.9651 & 0.9438 & 0.9277 & 0.9780\\
FP& 0.0595 & 0.0000 & 0.0348 & 0.0562 & 0.0723 & 0.0220\\
FN& 0.4375 & 0.5000 & 0.3571 & 0.3636 & 0.7059 & 0.3333\\ 
\bottomrule
\end{tabular}%
}
\end{table}

\subsection{Model Efficiency}
The choice of document representation and learning algorithm significantly affects the computational efficiency of emotion detection systems, as demonstrated by comparing the efficiency of time over five languages using FastText embeddings in Table~\ref{tab:time-efficiency}.
% The selection of document representation methodologies and learning algorithms can significantly influence the computational efficiency of the emotion detection systems, as evidenced by the empirical data in  embedding time analysis (Table \ref{tab:result-efficiency-embedding-time}), training time analysis (Table \ref{tab:result-efficiency-training-time}), and prediction time analysis (Table \ref{tab:result-efficiency-prediction-time}), training time and prediction time are evaluated after FastText embeddings. 
Simpler models, such as DT, exhibit rapid training speeds (0.50–1.14s), making them computationally efficient but often at the expense of accuracy. In contrast, MLP achieves superior predictive performance but requires significantly longer training times (24.84–53.01s), representing a substantial increase in computational cost over DT. The Voting classifier, which integrates multiple models, falls between these extremes, with training times ranging from 9.67s to 23.51s.
Despite these differences in training efficiency, all models achieve sub-millisecond inference speeds, with prediction times between 0.3 ms and 0.7 ms, except for Voting in Marathi (0.95 µs) and Russian (1.92 µs). This suggests that inference latency is primarily influenced by model architecture rather than language complexity.
These results highlight key efficiency-accuracy trade-offs. While high-dimensional embeddings like TF-IDF achieve strong F1 scores (e.g., Marathi: TF-IDF 0.68), their computational costs (140.24s) may be prohibitive for real-time applications. FastText with MLP provides a balanced alternative, offering competitive accuracy (Marathi: 0.67) with moderate computational cost (embedding: 1.00s, training: 43.56s), underscoring the need for multi-objective optimization in multilingual emotion detection.

\begin{table}
\caption{Train and test time in seconds using FastText embeddings.}
\label{tab:time-efficiency}
\centering
\resizebox{0.99\columnwidth}{!}{
\begin{tabular}{ccccccc}
\toprule
        \multirow{2}{*}{Language} & \multicolumn{2}{c}{DT} & \multicolumn{2}{c}{Voting} & \multicolumn{2}{c}{MLP} \\
        \cmidrule{2-7} 
        & Train & Test & Train & Test & Train & Test\\
\midrule
Marathi & 1.06 & 6e-4 & 23.51 & 9.54e-7 & 43.56 & 6e-4\\
Spanish & 0.86 & 4e-4 &17.22 & 0.0000 &34.39 & 7e-4\\
Hindi & 1.14  & 3e-4 &22.26 & 0.0000 &49.78 & 7e-4\\
Romanian & 0.50 & 3e-4 & 9.67& 0.0000 & 24.84 & 6e-4\\
Russian & 1.14 & 4e-4 & 22.69& 1.92e-6 & 53.01 & 7e-4\\
\bottomrule
\end{tabular}%
}
\end{table}

% \subsection{Future Work}

%future work 1: based on the feature and classifier combination, ensure that the most suitable emotion detection techniques are applied to each language.
%future work 2: Change the LLM model to the most up-to-date one to see whether they can find the most suitable language feature extraction model for various languages.
%future work 3: train fasttext word vectors：https://fasttext.cc/docs/en/unsupervised-tutorial.html
\section{Conclusions}
This study presents a feature-centric framework for cross-lingual multi-emotion detection in short texts, designed to  dynamically adapt of document representations and learning algorithms for optimal language-specific performance. 
Through a comprehensive comparative study across 28 languages—highlighting five for demonstration—we evaluate three key components: document representation, dimensionality reduction, and model training. Our findings show that the proposed pipeline is adaptable across languages with minimal adjustments, effectively balancing computational efficiency and detection accuracy. Experimental results validate its robustness in multi-label emotion prediction, particularly for low-resource languages.

In future work, we plan to refine the framework by optimizing feature-classifier selection for each language, leveraging advanced LLMs for enhanced feature extraction, and training FastText word vectors to improve representation quality, particularly for low-resource languages. We will also focus on determining optimal PCA configurations and evaluating its impact on the performance across different languages and emotion categories to ensure the robustness and reliability of our framework.

% The framework's flexibility addresses cross-lingual challenges through two mechanisms: (1) Language-agnostic feature engineering that enables the generation of document representations while ignoring linguistic diversity and preserves emotional semantics. (2) Adaptive learning algorithm selection that strikes a balance between computational efficiency and detection accuracy. 

% Experimental results validate the framework's effectiveness, showing its capability to perform robust and efficient cross-lingual multi-label emotion prediction, even for low-resource languages where traditional approaches fall short.
% Experimental results validate its effectiveness in analyzing different combinations of text representation methodologies and learning algorithms, demonstrating the system's ability to complete a robust and efficient cross-lingual multi-label emotion prediction task, even for tasks containing low-resource languages where traditional one-size-fits-all approaches do not work.

\section*{Ethical Statements}
This paper presents a feature-centric framework for cross-lingual multi-emotion detection, utilizing the publicly available BRIGHTER dataset \citep{muhammad2025brighterbridginggaphumanannotated}. While leveraging its multilingual resources, we explicitly acknowledge that emotional expression is culturally and linguistically dependent, which may introduce biases in annotation and model predictions, particularly in capturing nuanced emotional expressions across low-resource languages and dialects. To address these challenges, we advocate for responsible development and deployment of the framework, emphasizing ongoing research into bias detection, fairness in cross-lingual emotion analysis, and mitigation of potential data-driven biases. These efforts aim to ensure equitable and ethical applications of the technology while transparently addressing its cultural and linguistic limitations.

\section*{Acknowledgments}
We would like to thank the organizers of SemEval 2025 Task 11 for hosting the shared task and providing valuable resources that enabled this research. Their efforts in promoting multilingual emotion analysis are greatly appreciated. We are also grateful to the anonymous reviewers for their insightful feedback and constructive suggestions, which have helped us improve the quality and clarity of this work.

\bibliography{custom}

\clearpage

\appendix
\section{LLM query for Language Family Classification}
\label{sec:appendix-land-class}

The LLM employed in our system is Baidu \texttt{Qianfan} "ernie-4.0-8k-latest", and the prompt is "\texttt{You are a linguist working on language classification and are familiar with the given languages: (list the known languages). Please select the language from the list that is most similar to (given language) based on language family and geographic distance in terms of population distribution.}".

\section{Data Statics}\label{sec:appendix-data-statistics}
Table~\ref{tab:appendix-stats-table} presents the number of training, development and test splits across all 28 languages in our experiments, sorted by number of training samples in descending order.

% \usepackage{graphicx}
\begin{table}[H]
\centering
\caption{Train (train), Development (dev) and Test (test) splits for all 28 languages.}
\label{tab:appendix-stats-table}
\resizebox{0.95\columnwidth}{!}{%
\begin{tabular}{llll}
\toprule
\multicolumn{1}{l}{\textbf{Language}} & \multicolumn{1}{c}{\textbf{\#train}} & \multicolumn{1}{c}{\textbf{\#dev}} & \multicolumn{1}{c}{\textbf{\#test}} \\
\midrule
Nigerian pidgin                            & 3728                                       & 620                                      & 1870                                      \\
Tigrinya                                   & 3681                                       & 614                                      & 1840                                      \\
Amharic                                    & 3549                                       & 592                                      & 1774                                      \\
Oromo                                      & 3442                                       & 574                                      & 1721                                      \\
Somali                                     & 3392                                       & 566                                      & 1696                                      \\
Swahili                                    & 3307                                       & 551                                      & 1656                                      \\
Yoruba                                     & 2992                                       & 497                                      & 1500                                      \\
Igbo                                       & 2880                                       & 479                                      & 1444                                      \\
English                                    & 2768                                       & 116                                      & 2767                                      \\
Russian                                    & 2679                                       & 199                                      & 1000                                      \\
Chinese                                    & 2642                                       & 200                                      & 2642                                      \\
German                                     & 2603                                       & 200                                      & 2604                                      \\
Hindi                                      & 2556                                       & 100                                      & 1010                                      \\
Ukrainian                                  & 2466                                       & 249                                      & 2234                                      \\
Kinyarwanda                                & 2451                                       & 407                                      & 1231                                      \\
Marathi                                    & 2415                                       & 100                                      & 1000                                      \\
Portuguese(Brazilian)                      & 2226                                       & 200                                      & 2226                                      \\
Hausa                                      & 2145                                       & 356                                      & 1080                                      \\
Spanish                                    & 1996                                       & 184                                      & 1695                                      \\
Moroccan Arabic                            & 1608                                       & 267                                      & 812                                       \\
Makhuwa                                    & 1551                                       & 258                                      & 777                                       \\
Portuguese(Mozambican)                     & 1546                                       & 257                                      & 776                                       \\
Romanian                                   & 1241                                       & 123                                      & 1119                                      \\
Afrikaans                                  & 1222                                       & 98                                       & 1065                                      \\
Swedish                                    & 1187                                       & 200                                      & 1188                                      \\
Tatar                                      & 1000                                       & 200                                      & 1000                                      \\
Sundanese                                  & 924                                        & 199                                      & 926                                       \\
Algerian Arabic                            & 901                                        & 100                                      & 902                                             
\\
\bottomrule
\end{tabular}%
}
\end{table}

\newpage
% \clearpage
\section{Model Performance}\label{sec:appendix-prediction-accuracy}
Table~\ref{tab:appendix-all-scores} shows the F1-macro scores across all 28 languages using three representation methods (TF-IDF, FastText and SBERT) combined with three classifiers (DT, Voting and MLP). The languages are listed in descending order based on the number of training samples, aligning with the organizational schema of Appendix~\ref{sec:appendix-data-statistics}.

Overall, the transformer-based document representations, such as Sentence-BERT, generally outperform TF-IDF and FastText across 18 out of 28 languages, demonstrating their effectiveness in capturing semantic nuances. In contrast, the pretrained word embeddings such as FastText tend to yields lower scores in most cases, likely due to the limited representation of several low-resource or less commonly used languages in its pretraining corpus, resulting in suboptimal embedding quality for these languages. On the classifier side, the deep learning model MLP consistently delivers the best performance, particularly when combined with SBERT representation, highlighting the advantage of deep learning models in leveraging dense contextual representations. Traditional machine learning approaches, such as Decision Tree, performs better with sparse features like TF-IDF but struggles with dense representations. These findings collectively underscore the importance of combining semantically rich representations with expressive classifiers for robust multilingual emotion detection.

\begin{table}[]
\centering
\caption{F1-macro scores for all 28 languages.}
\label{tab:appendix-all-scores}
\resizebox{\textwidth}{!}{
\begin{tabular}{c|lllllc|llll}
\cline{1-5} \cline{7-11}
\multicolumn{1}{l|}{\textbf{Language}} &  & \multicolumn{1}{c}{\textbf{DT}} & \multicolumn{1}{c}{\textbf{Voting}} & \multicolumn{1}{c}{\textbf{MLP}} &  & \multicolumn{1}{l|}{\textbf{Language}} &  & \multicolumn{1}{c}{\textbf{DT}} & \multicolumn{1}{c}{\textbf{Voting}} & \multicolumn{1}{c}{MLP} \\ \cline{1-5} \cline{7-11}
\multirow{3}{*}{\begin{tabular}[c]{@{}c@{}}Nigerian \\ pidgin\end{tabular}} & TF-IDF & 0.3268 & 0.2764 & 0.3457 &  & \multirow{3}{*}{Kinyarwanda} & TF-IDF & 0.2779 & 0.2595 & 0.3163 \\
 & FastText & 0.2681 & 0.2600 & 0.3990 &  &  & FastText & 0.2272 & 0.1815 & 0.1751 \\
 & SBERT & 0.3065 & 0.2721 & 0.4142 &  &  & SBERT & 0.1834 & 0.2146 & 0.3432 \\ \cline{1-5} \cline{7-11}
\multirow{3}{*}{Tigrinya} & TF-IDF & 0.2473 & 0.1758 & 0.2846 &  & \multirow{3}{*}{Marathi} & TF-IDF & 0.6817 & 0.7438 & 0.7640 \\
 & FastText & 0.0477 & 0.0430 & 0.0208 &  &  & FastText & 0.3849 & 0.4277 & 0.6711 \\
 & SBERT & 0.1957 & 0.1791 & 0.1969 &  &  & SBERT & 0.4275 & 0.6654 & 0.8389 \\ \cline{1-5} \cline{7-11}
\multirow{3}{*}{Amharic} & TF-IDF & 0.2281 & 0.2557 & 0.3418 &  & \multirow{3}{*}{\begin{tabular}[c]{@{}c@{}}Portuguese\\ (Brazilian)\end{tabular}} & TF-IDF & 0.2170 & 0.1723 & 0.2296 \\
 & FastText & 0.0569 & 0.1216 & 0.0142 &  &  & FastText & 0.2004 & 0.1648 & 0.3198 \\
 & SBERT & 0.3014 & 0.3111 & 0.4121 &  &  & SBERT & 0.2821 & 0.2681 & 0.5131 \\ \cline{1-5} \cline{7-11}
\multirow{3}{*}{Oromo} & TF-IDF & 0.3222 & 0.3351 & 0.4172 &  & \multirow{3}{*}{Hausa} & TF-IDF & 0.4355 & 0.5250 & 0.5560 \\
 & FastText & 0.1756 & 0.2046 & 0.1033 &  &  & FastText & 0.3225 & 0.3250 & 0.3538 \\
 & SBERT & 0.1974 & 0.1969 & 0.2397 &  &  & SBERT & 0.3065 & 0.3445 & 0.4667 \\ \cline{1-5} \cline{7-11}
\multirow{3}{*}{Somali} & TF-IDF & 0.2481 & 0.2704 & 0.3523 &  & \multirow{3}{*}{Spanish} & TF-IDF & 0.5516 & 0.6561 & 0.6284 \\
 & FastText & 0.1372 & 0.0756 & 0.1296 &  &  & FastText & 0.3825 & 0.4046 & 0.6479 \\
 & SBERT & 0.1460 & 0.1408 & 0.2577 &  &  & SBERT & 0.5022 & 0.5867 & 0.7076 \\ \cline{1-5} \cline{7-11}
\multirow{3}{*}{Swahili} & TF-IDF & 0.2015 & 0.1352 & 0.1966 &  & \multirow{3}{*}{\begin{tabular}[c]{@{}c@{}}Moroccan \\ Arabic\end{tabular}} & TF-IDF & 0.2142 & 0.2014 & 0.2838 \\
 & FastText & 0.1563 & 0.0653 & 0.1710 &  &  & FastText & 0.1942 & 0.2182 & 0.3748 \\
 & SBERT & 0.1469 & 0.0913 & 0.1442 &  &  & SBERT & 0.2530 & 0.3409 & 0.4183 \\ \cline{1-5} \cline{7-11}
\multirow{3}{*}{Yoruba} & TF-IDF & 0.2056 & 0.2085 & 0.2610 &  & \multirow{3}{*}{Makhuwa} & TF-IDF & 0.2055 & 0.1125 & 0.1554 \\
 & FastText & 0.1495 & 0.1279 & 0.1899 &  &  & FastText & 0.0869 & 0.0393 & 0.0640 \\
 & SBERT & 0.1432 & 0.0782 & 0.1989 &  &  & SBERT & 0.1366 & 0.0556 & 0.0985 \\ \cline{1-5} \cline{7-11}
\multirow{3}{*}{Igbo} & TF-IDF & 0.3789 & 0.4140 & 0.4888 &  & \multirow{3}{*}{\begin{tabular}[c]{@{}c@{}}Portuguese\\ (Mozambican)\end{tabular}} & TF-IDF & 0.2651 & 0.1977 & 0.1571 \\
 & FastText & 0.2321 & 0.2867 & 0.3753 &  &  & FastText & 0.1514 & 0.1009 & 0.3009 \\
 & SBERT & 0.2591 & 0.2809 & 0.3671 &  &  & SBERT & 0.1794 & 0.2147 & 0.4021 \\ \cline{1-5} \cline{7-11}
\multirow{3}{*}{English} & TF-IDF & 0.3302 & 0.3908 & 0.5197 &  & \multirow{3}{*}{Romanian} & TF-IDF & 0.4484 & 0.4358 & 0.6110 \\
 & FastText & 0.4177 & 0.4069 & 0.6139 &  &  & FastText & 0.4633 & 0.4486 & 0.6217 \\
 & SBERT & 0.4421 & 0.5683 & 0.6654 &  &  & SBERT & 0.5167 & 0.5630 & 0.6375 \\ \cline{1-5} \cline{7-11}
\multirow{3}{*}{Russian} & TF-IDF & 0.6418 & 0.7107 & 0.7155 &  & \multirow{3}{*}{Afrikaans} & TF-IDF & 0.2153 & 0.2366 & 0.1813 \\
 & FastText & 0.3202 & 0.3472 & 0.6341 &  &  & FastText & 0.1763 & 0.1155 & 0.1283 \\
 & SBERT & 0.4661 & 0.5767 & 0.7188 &  &  & SBERT & 0.2655 & 0.2834 & 0.4729 \\ \cline{1-5} \cline{7-11}
\multirow{3}{*}{Chinese} & TF-IDF & 0.2730 & 0.2964 & 0.3889 &  & \multirow{3}{*}{Swedish} & TF-IDF & 0.3415 & 0.2694 & 0.2542 \\
 & FastText & 0.0904 & 0.1207 & 0.0711 &  &  & FastText & 0.2619 & 0.2715 & 0.3729 \\
 & SBERT & 0.3475 & 0.3831 & 0.5402 &  &  & SBERT & 0.3144 & 0.3118 & 0.4310 \\ \cline{1-5} \cline{7-11}
\multirow{3}{*}{German} & TF-IDF & 0.3054 & 0.2889 & 0.3611 &  & \multirow{3}{*}{Tatar} & TF-IDF & 0.4457 & 0.4724 & 0.4528 \\
 & FastText & 0.3116 & 0.3101 & 0.4064 &  &  & FastText & 0.2508 & 0.2339 & 0.3885 \\
 & SBERT & 0.3130 & 0.3535 & 0.5011 &  &  & SBERT & 0.2418 & 0.2385 & 0.3777 \\ \cline{1-5} \cline{7-11}
\multirow{3}{*}{Hindi} & TF-IDF & 0.5092 & 0.4927 & 0.6122 &  & \multirow{3}{*}{Sundanese} & TF-IDF & 0.3503 & 0.3642 & 0.3690 \\
 & FastText & 0.3261 & 0.3067 & 0.6051 &  &  & FastText & 0.2455 & 0.2104 & 0.2452 \\
 & SBERT & 0.4490 & 0.5682 & 0.7374 &  &  & SBERT & 0.2407 & 0.2851 & 0.3605 \\ \cline{1-5} \cline{7-11}
\multirow{3}{*}{Ukrainian} & TF-IDF & 0.2751 & 0.2906 & 0.2735 &  & \multirow{3}{*}{\begin{tabular}[c]{@{}c@{}}Algerian \\ Arabic\end{tabular}} & TF-IDF & 0.3464 & 0.3311 & 0.4770 \\
 & FastText & 0.1384 & 0.1123 & 0.2092 &  &  & FastText & 0.3109 & 0.3343 & 0.4421 \\
 & SBERT & 0.2675 & 0.2874 & 0.4472 &  &  & SBERT & 0.3642 & 0.3782 & 0.5126 \\ \cline{1-5} \cline{7-11}
\end{tabular}
}
\end{table}

\end{document}